\documentclass[runningheads]{llncs}
\usepackage[T1]{fontenc}
\usepackage{graphicx}
\usepackage{amsmath}   
\usepackage{amssymb}   
\usepackage{booktabs}  
\usepackage{multirow}   
\usepackage{hyperref}
\usepackage{marvosym}  
\hypersetup{colorlinks=true, linkcolor=blue, citecolor=blue, urlcolor=blue}
\begin{document}
\title{Teacher Encoder-Student Decoder Denoising Guided Segmentation Network for Anomaly Detection}
\titlerunning{Teacher-Encoder Student-Decoder Network for Anomaly Detection}
%
\author{Shixuan Song\inst{1} \and
	Hao Chen\inst{1}\textsuperscript{(\Letter)} \and
	Shu Hu\inst{2} \and
	Xin Wang\inst{3} \and
	Jinrong Hu\inst{1} \and
	Xi Wu\inst{1}}
\authorrunning{S. Song et al.}
%
\institute{School of Computer Science, Chengdu University of Information Science and Technology, Chengdu 610225, China\\
	\email{3230604024@stu.cuit.edu.cn},
	\email{\{haochen,hjr,wuxi\}@cuit.edu.cn} \and
	Purdue University, West Lafayette 47907, USA\\
	\email{hu968@purdue.edu} \and
	University at Albany, NY, USA\\
	\email{xwang56@albany.edu}}
\maketitle              
\begin{abstract}
	Visual anomaly detection is a highly challenging task, often categorized as a one-class classification and segmentation problem. Recent studies have demonstrated that the student-teacher (S-T) framework effectively addresses this challenge. However, most S-T frameworks rely solely on pre-trained teacher networks to guide student networks in learning multi-scale similar features, overlooking the potential of the student networks to enhance learning through multi-scale feature fusion. In this study, we propose a novel model named PFADSeg, which integrates a pre-trained teacher network, a denoising student network with multi-scale feature fusion, and a guided anomaly segmentation network into a unified framework. By adopting a unique teacher-encoder and student-decoder denoising mode, the model improves the student network's ability to learn from teacher network features. Furthermore, an adaptive feature fusion mechanism is introduced to train a self-supervised segmentation network that synthesizes anomaly masks autonomously, significantly increasing detection performance. Rigorous evaluations on the widely-used MVTec AD dataset demonstrate that PFADSeg exhibits excellent performance, achieving an image-level AUC of 98.9\%, a pixel-level mean precision of 76.4\%, and an instance-level mean precision of 78.7\%. 
	
\end{abstract}
\section{Introduction}
The primary objective of anomaly detection is not only to identify anomalous images but also to precisely segment the anomalous pixel regions. It is critical in various computer vision applications, including industrial defect inspection \cite{Roth2021TowardsTR,zavrtanik2021draem}, medical image analysis \cite{schlegl2017unsupervised}, and video surveillance \cite{Liu2018ClassifierTS}. However, these tasks commonly face significant challenges, such as the scarcity of anomalous samples and the diversity of anomaly types. This variability makes it difficult to collect sufficient data or labels for the wide range of anomaly categories. Furthermore, the differences between normal and anomalous samples are frequently subtle, often masked by irrelevant noise, which complicates the task of simultaneously filtering out noise and accurately identifying anomalous pixel regions.
\begin{figure}
	\centering
	\includegraphics[width=0.6\textwidth]{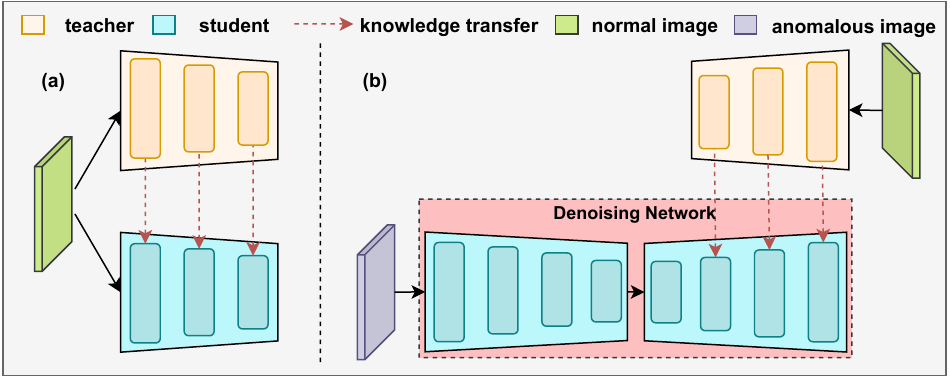}
	\caption{The traditional S-T framework (a) and our denoising student-teacher distillation framework (b) in the KD mode.}
	\label{fig:Distinction}
\end{figure}

Due to the scarcity of anomaly data and labeling difficulties, anomaly detection often employs unsupervised methods, training models exclusively on normal data. Recently, the knowledge distillation mode \cite{Salehi2020MultiresolutionKD}, particularly the student-teacher (S-T) framework, has demonstrated its effectiveness in anomaly detection tasks \cite{bergmann2020uninformed,deng2022anomaly,Salehi2020MultiresolutionKD,Wang2021StudentTeacherFP,Yamada2021ReconstructionSW}. In this framework, a teacher network, typically pre-trained on large datasets (e.g., ImageNet \cite{deng2009imagenet}) with well-established weights, guides a student network. The student, often sharing the teacher's architecture, is trained on anomaly detection datasets \cite{bergmann2019mvtec} containing only normal samples, aiming to mimic the teacher's feature representations for these samples. Furthermore, knowledge distillation across multiple feature pyramid levels \cite{Salehi2020MultiresolutionKD,Wang2021StudentTeacherFP,Zhang2022DeSTSegSG} allows for the integration of multi-level feature differences, often leading to enhanced performance. Despite these advancements, several challenges persist. First, the lack of anomalous samples during student network training may lead to insufficient representation of feature differences between the teacher and student networks for anomalies in the S-T framework. Second, the student network in most S-T frameworks shares the same architecture as the teacher network, which may limit the student's ability to effectively learn features extracted by the teacher, particularly in large-scale anomaly detection datasets where training the student network becomes more complex.

\begin{figure}[t]
	\centering
	\includegraphics[width=\textwidth]{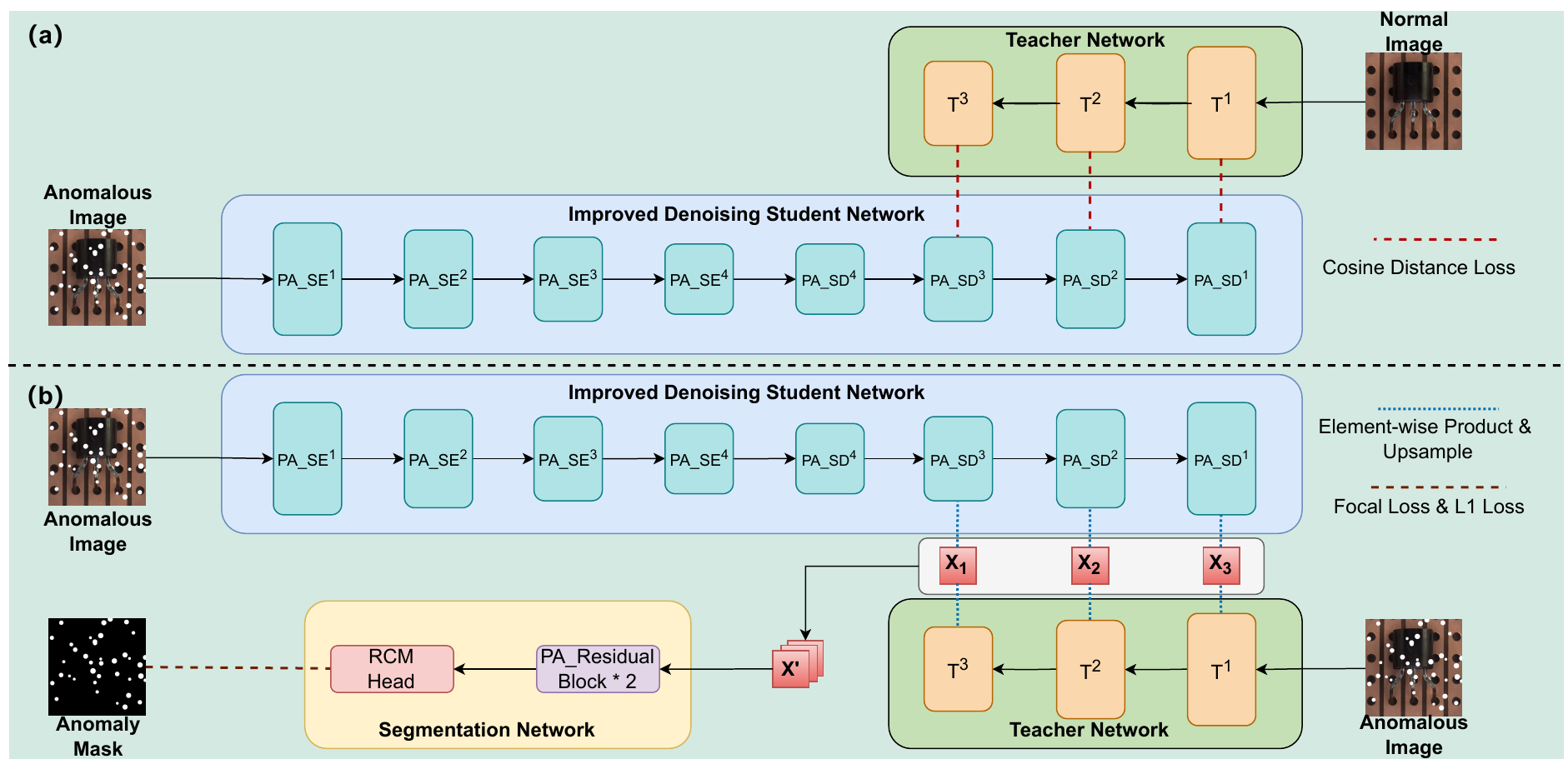}
	\caption{Overview of PFADSeg. The training consists of two stages. In stage (a), the improved student network is trained with pseudo-anomalous images (created by blending anomaly masks with normal images), while the teacher network, with fixed weights, processes normal images. The student learns to match features extracted by the teacher. In stage (b), both networks take pseudo-anomalous images as input, with weights fixed. Their normalized outputs are combined via element-wise multiplication and concatenation to form input for the segmentation network, which is trained using the anomaly masks as ground truth.}
	\label{fig:PFADSeg}
\end{figure}

To address the aforementioned challenges, we introduce a novel knowledge distillation (KD) approach, the denoising student-teacher distillation mode (see Fig.~\ref{fig:Distinction}), where paired normal and anomalous samples are fed into the teacher and student networks, respectively, with the student network adopting a different architecture from the teacher. Based on this mode, we propose a Parallel Feature Aggregation and Denoising for Anomaly Segmentation (PFADSeg, Fig.~\ref{fig:PFADSeg}), which consists of a pre-trained teacher network, an improved denoising student network, and a segmentation network. Synthetic anomaly masks are added to normal images to generate inputs for the student network, while the teacher processes the corresponding unaltered normal images. This setup allows the student network to remove noise and learn features aligned with those of the teacher network, leveraging architectural differences to reduce sensitivity to irrelevant noise. The student network, based on ResNet18 \cite{he2016deep}, is further enhanced to better fuse features across multiple levels. Finally, a trainable segmentation network is introduced to process multi-level feature differences between the teacher and student networks, with synthetic anomaly masks providing supervision during training.

We evaluated the performance of our model using the unsupervised MVTec AD dataset \cite{bergmann2019mvtec}. Extensive experimental results demonstrate that our approach achieves significant improvements in anomaly detection tasks at the image level, pixel level, and instance level. Additionally, we conducted ablation studies to validate the effectiveness of our proposed model enhancements. Compared to previous work \cite{Zhang2022DeSTSegSG}, our method introduces several improvements to the segmentation network and the denoising student network:
\begin{itemize}
	\item We enhance the segmentation network's ability to capture global contextual information by incorporating horizontal and vertical pooling. Additionally, we introduce a large-kernel strip convolution to create rectangular attention regions, which better highlight anomalous pixels.
	
	\item We improve the denoising student network by modifying the residual block structure based on ResNet18 \cite{he2016deep}. Specifically, we introduce an attention mechanism into the residual connections, enabling dynamic feature fusion and allowing the network to simultaneously focus on both globally distributed large objects and locally distributed small objects.
	
	\item We propose a Parallel Convolutional Attention Recalibration Module to replace the second convolutional layer in the residual blocks, which further enhances feature extraction and anomaly detection performance.
\end{itemize}

\section{Related Works}
Anomaly detection and localization have been explored from various perspectives, with significant progress in methods based on image reconstruction, memory banks, and knowledge distillation.

In image reconstruction, common models include autoencoders \cite{bergmann2018improving}, variational autoencoders (VAEs) \cite{baur2019deep}, and generative adversarial networks (GANs) \cite{Schlegl2019fAnoGANFU,schlegl2017unsupervised}. Trained on normal data, these models aim to poorly reconstruct unseen anomalous images. The pixel-wise difference between the input image and the reconstructed image is used as an anomaly score. However, a key challenge is overgeneralization, where anomalous regions might be accurately reconstructed, diminishing the anomaly score's effectiveness \cite{pirnay2022inpainting}. To mitigate this, recent works incorporate adversarial training or hybrid models that combine reconstruction with discriminative learning for improved anomaly robustness.

Memory-based methods \cite{Roth2021TowardsTR,Yi2020PatchSP} construct a memory bank from normal training data. During inference, anomaly scores are derived from the similarity between a query and its closest matches in the memory bank. These methods effectively capture fine-grained patterns in normal data, especially for subtle or context-dependent anomalies. Extensions incorporating attention mechanisms or dynamic memory updates further enhance their real-world applicability.

Knowledge distillation, leveraging a pre-trained teacher network and a trainable student network, is highly effective for anomaly detection. The student network is typically trained on a dataset consisting of only normal data, and its anomalous feature representations are expected to differ from those of the teacher network. This divergence provides a basis for distinguishing anomalies. Various strategies have been proposed to enhance the detection of different anomaly types. For instance, Bergmann et al. \cite{bergmann2020uninformed} employed ensemble learning with multiple students, leveraging irregularities in their feature representations for more effective anomaly detection. Salehi et al. \cite{Salehi2020MultiresolutionKD} and Wang et al. \cite{Wang2021StudentTeacherFP} introduced multi-level feature alignment to capture diverse anomalies. These methods demonstrate the versatility and adaptability of knowledge distillation in addressing various challenges in anomaly detection. To prevent identical architectures and inputs for teacher and student networks, some works \cite{deng2022anomaly,Yamada2021ReconstructionSW} utilize a student-decoder architecture. Furthermore, Zhang et al. \cite{Zhang2022DeSTSegSG} further distinguishes the representations of anomalies between the teacher and student networks within the S-T framework.

Anomaly Simulation. For one-class anomaly detection (AD) where training anomalies are absent, simulating pseudo-anomalous data enables supervised model training. Traditional simulation methods, like rotation and cropping \cite{schlegl2017unsupervised}, often prove ineffective for fine-grained anomaly patterns. Zhang et al. \cite{Zhang2022DeSTSegSG} adopted the idea proposed in \cite{zavrtanik2021draem} , using two-dimensional Perlin noise to simulate more realistic anomalous images, and utilized the generated anomaly masks as ground truth for the segmentation network. However, such an approach risks the segmentation network overlooking the impact of irrelevant noise in normal images when segmenting actual anomalies. This can lead to misclassifying normal regions as anomalous during inference. Focusing on irrelevant noise can detract from true anomalies, reducing detection effectiveness, particularly for certain texture categories in datasets like MVTec AD \cite{bergmann2019mvtec}.

\section{Method}
The proposed PFADSeg architecture consists of three main components: a teacher network pre-trained on the ImageNet \cite{deng2009imagenet} dataset, an improved denoising student network, and a segmentation network. The training process utilizes synthetic anomaly images generated based on the algorithm in \cite{zavrtanik2021draem}. Specifically, random 2D Perlin noise is binarized to create an anomaly mask \(k\). The masked regions are replaced with a linear combination of the normal image \(M_{n}\) and an external data source \(A\), producing the synthetic anomaly image \(M_{a}\). The transparency factor $\beta$ is sampled from \([0.15,1]\), and \(\odot\) represents element-wise multiplication. The process is defined as follows:
\begin{equation}
	M_a = \beta(K \odot A) + (1 - \beta)(K \odot M_n) + (1 - K) \odot M_n \label{eq:1}
\end{equation}

As illustrated in Fig.~\ref{fig:PFADSeg}, the training is carried out in two steps: in the first step, the synthetic anomaly images are used as inputs to the student network, while the corresponding original images (without anomaly masks) are used as inputs to the teacher network. The weights of the improved denoising student network are trainable, whereas the teacher network’s weights remain fixed. The goal of this stage is to enable the student network to denoise the input while ensuring that the features extracted by its decoder align as closely as possible with the features extracted by the teacher network. In the second step, The student network's weights are frozen, and both networks process the synthetic anomaly images as input. The features extracted by the teacher and student networks are fed into the segmentation network, which is trained using the synthetic anomaly masks as ground truth in a self-supervised manner.

\subsection{Teacher Encoder-Student Decoder Denoising Network}
A key aspect of our approach is the distinct architecture employed by the denoising student network compared to the teacher. The teacher network is based on a ResNet18 \cite{he2016deep} pre-trained on ImageNet, utilizing only its first three convolutional blocks (conv1\_x, conv2\_x, conv3\_x), denoted \(T_1\), \(T_2\), and \(T_3\), respectively. In contrast, the denoising student network features an encoder based on an enhanced ResNet18 architecture (termed PA\_ResNet18), featuring modified residual blocks. This encoder comprises four convolutional blocks (\(PA\_SE^1\) to \(PA\_SE^4\)). The student's decoder is constructed similarly but in reverse (replacing downsampling with bilinear upsampling) and also consists of four convolutional blocks (\(PA\_SD^1\) to \(PA\_SD^4\)).

Feature fusion, the combination of features from different layers or branches, commonly uses simple operations like summation or concatenation. However, such simple operations can be suboptimal. Seeking more effective fusion, we draw inspiration from the Attentional Feature Fusion (AFF) module \cite{dai2021attentional}. Recognizing that anomaly detection requires precise region segmentation, not just binary classification, our goal is to develop a feature fusion mechanism suitable for diverse anomaly scenarios. Our approach incorporates attention mechanisms and multi-scale context aggregation to effectively gather contextual information across different receptive fields. This enables the network to adapt to objects of different scales, focusing on both large objects with global distribution and small objects with local distribution. Such adaptability is crucial for anomaly detection, particularly for segmenting small anomalous regions accurately.
\begin{figure}[b]
	\centering
	\includegraphics[width=0.5\textwidth]{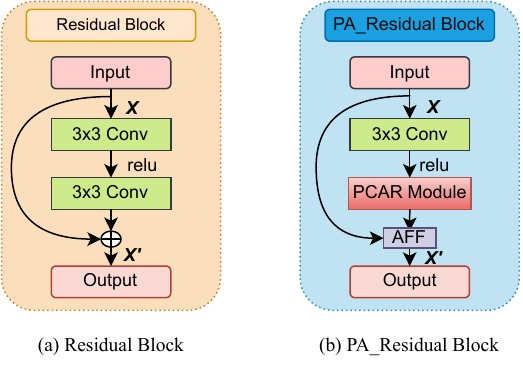}
	\caption{Comparison between the original Residual Block (a) in ResNet18 and the proposed PA\_Residual Block (b).}
	\label{fig:PA_Residual}
\end{figure}

The architecture of the enhanced residual block used in our PA\_ResNet18 is shown in Fig.~\ref{fig:PA_Residual}. Instead of the standard summation in the residual connection, we employ the Attentional Feature Fusion (AFF) mechanism for dynamic feature fusion. Furthermore, the second \(3 \times 3\) convolution layer is replaced by our proposed Parallel Convolutional Attention Recalibration (PCAR) module (detailed in Sec.~\ref{3.2}). These modifications enhance the block's ability to focus on relevant features, thereby improving overall anomaly detection and segmentation performance.

The training objective is to minimize the cosine distance between corresponding feature maps from the teacher blocks (\(T_i\)) and the student decoder blocks (\(PA\_SD^i\)) for \(i = 1, 2, 3\). Let \( F_{T_i} \in \mathbb{R}^{C_i \times H_i \times W_i} \) and \( F_{S_i} \in \mathbb{R}^{C_i \times H_i \times W_i} \) denote these feature maps, respectively. We first compute the cosine similarity \(X_i(j,k)\) and subsequently the cosine distance \(D_i(j,k)\) between the feature vectors at each spatial location \((j, k)\) (where \( 1 \le j \le H_i, 1 \le k \le W_i \)), following Equations \eqref{eq:2} and \eqref{eq:3}. The overall cosine distance loss, \( L_{\text{cos}} \), guiding the training, is the sum of these distances across all locations and all three feature pairs (\(i=1,2,3\)), as formulated in Equation \eqref{eq:4}.
\begin{equation}
	X_i(j, k) = \frac{F_{T_i}(j, k) \odot F_{S_i}(j, k)}{\left\|F_{T_i}(j, k)\right\|_2 \left\|F_{S_i}(j, k)\right\|_2}
	\label{eq:2}
\end{equation}

\begin{equation}
	D_i(j, k) = 1 - \sum_{c=1}^{C_i} X_i(j, k)_c
	\label{eq:3}
\end{equation}

\begin{equation}
	L_{\cos} = \sum_{i=1}^{3} \left( \frac{1}{H_i W_i} \sum_{j,k=1}^{H_i,W_i}  D_i(j, k) \right)
	\label{eq:4}
\end{equation}

\subsection{Parallel Convolutional Attention Recalibration Module}\label{3.2}
\begin{figure}[t]
	\centering
	\includegraphics[width=1.0\textwidth]{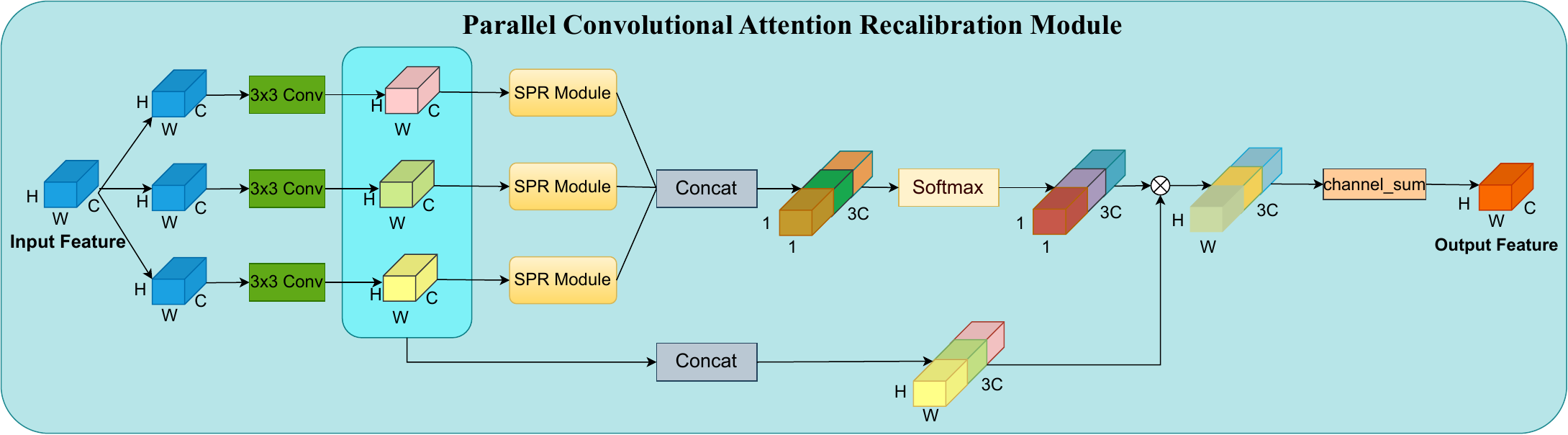}
	\caption{Overall architecture of the PCAR module. The feature map is represented by its dimensions \( H \), \( W \), and \( C \), denoting height, width, and the number of channels, respectively. The Softmax operation recalibrates channel attention weights to capture long-range channel dependencies. The symbol \( \otimes \) indicates element-wise multiplication, and channel\_sum refers to summation along the channel dimension.}
	\label{fig:PCAR}
\end{figure}

ResNet18 employs residual blocks within its convolutional layers (i.e., conv2\_x, conv3\_x, conv4\_x, conv5\_x). We propose a novel Parallel Convolutional Attention Recalibration (PCAR) Module to enhance these blocks. Employing parallel convolutions to extract diverse spatial information, the PCAR module aims to improve the localization and segmentation of anomalous regions while suppressing irrelevant noise. Specifically, PCAR replaces the second \(3 \times 3\) convolution layer within the standard ResNet18 residual block. The detailed architecture of the PCAR module is illustrated in Fig.~\ref{fig:PCAR}.

The PCAR module takes the feature map \( F \in \mathbb{R}^{H \times W \times C} \) (output from the block's first \(3 \times 3\) convolution) as input. Here, \(H, W, C\) denote height, width, and channels, respectively. This input \(F\) is processed by three parallel \(3 \times 3\) convolutional layers. Each path independently extracts features (\(F_i\), for \(i=1, 2, 3\)), thus enhancing the diversity of the features. This operation is defined as:
\begin{equation}
	F_i = \text{Conv}{3 \times 3}(F)
	\label{eq:5}
\end{equation}

Each resulting feature map \(F_i\) is then processed by a Spatial Pyramid Recalibration (SPR) module \cite{Yu2024MultiscaleSP}, designed to reinforce salient information within \(F_i\). The SPR module helps focus on critical feature regions while suppressing noise, thus enhancing segmentation precision. The SPR-processed feature maps (\(F_i'\)) are subsequently concatenated along the channel dimension:
\begin{equation}
	F' = \text{Concat}(F_1', F_2', F_3')
	\label{eq:6}
\end{equation}

Channel weights \( W \in \mathbb{R}^{1 \times 1 \times 3C} \) are generated from \(F'\) via Softmax activation:
\begin{equation}
	W = \text{Softmax}(F')
	\label{eq:7}
\end{equation}

The generated weight map \(W\) is then element-wise multiplied with the concatenated feature map \( F_1\), \(F_2\), \(F_3\) yielding the recalibrated feature map \( T \in \mathbb{R}^{H \times W \times 3C} \). This operation can be expressed as:
\begin{equation}
	T = W \odot \text{Concat}(F_1, F_2, F_3)
	\label{eq:8}
\end{equation}

Finally, a channel reduction operation is applied to \(T\) to restore the original channel dimension \(C\), producing the final output feature map \( Y \in \mathbb{R}^{H \times W \times C} \):
\begin{equation}
	Y = \text{channel\_sum}(T)
	\label{eq:9}
\end{equation}

In experimental results on the MVTec AD anomaly detection dataset, the newly proposed PCAR module demonstrated significant performance advantages. Thanks to its innovative parallel convolution design, PCAR is able to capture image features from multiple perspectives simultaneously, which not only enhances the model’s ability to detect subtle anomalies but also improves the accuracy of anomaly detection. The attention mechanism integrated within the PCAR module enables the model to adaptively focus on key areas of the image, ensuring accurate anomaly detection even in complex backgrounds, thus enhancing the model's robustness.

\subsection{Segmentation Network}
In the works of \cite{Salehi2020MultiresolutionKD}, anomaly scores are typically represented by directly summing the cosine distances of multi-layer features. However, if the discriminative abilities of the features across layers vary, this direct summation may result in suboptimal performance. To address this issue, we introduce a segmentation network that provides additional supervisory signals to guide the feature fusion process. Specifically, we freeze the weights of both the student and teacher networks to train the segmentation network. The synthetic anomalous images are passed as input to both the teacher-student network, while the corresponding binary anomaly masks are used as ground truth during training.

For input to the segmentation network, we compute the cosine similarity maps \(X_1, X_2, X_3\) between the paired features \((T_1, PA\_SD^1)\), \((T_2, PA\_SD^2)\), \((T_3, PA\_SD^3)\) using Equation \eqref{eq:2}. These similarity maps are then upsampled to match the spatial dimensions of \(X_1\). The upsampled maps, denoted \(\widehat{X}_1, \widehat{X}_2, \widehat{X}_3\), are concatenated along the channel dimension to form the input \(\widehat{X}\) for the segmentation network. Our segmentation network primarily consists of two enhanced PA\_Residual Blocks and a Rectangular Self-Calibration Module (RCM) \cite{Ni2024ContextGuidedSF}. While prior work like \cite{Zhang2022DeSTSegSG} used Atrous Spatial Pyramid Pooling (ASPP) for multi-scale context, its dilated convolutions can overlook small objects, leading to imprecise localization and over-segmentation, especially for the small anomalies common in datasets like MVTec AD \cite{bergmann2019mvtec}. Therefore, we replace ASPP with RCM \cite{Ni2024ContextGuidedSF}. RCM utilizes adjustable rectangular attention regions, enabling better focus on foreground anomalies of varying shapes and improving localization precision. This helps accurately delineate anomaly boundaries while minimizing the inclusion of non-anomalous areas.

To optimize the segmentation network's training, we employ a combination strategy of focal loss and L1 loss. Since most pixels in the training dataset belong to the easily recognized normal background, while anomalous pixels constitute a small yet critical minority, focal loss is used to emphasize these minority and hard-to-distinguish samples, improving the model's ability to segment anomalies. L1 loss is further introduced to enhance output sparsity, ensuring clearer segmentation boundaries. The ground truth anomaly mask is downsampled to a quarter of the input image size to match the output dimensions \((H_1, W_1)\). Let \(\widehat{Y}\) denote the output probability map and \(K\) the downsampled anomaly mask; the focal loss is computed as follows:
\begin{equation}
	L_{focal} = -\frac{1}{H_1 W_1} \sum_{i, j=1}^{H_1, W_1} (1 - q_{ij})^\gamma \log(q_{ij})
	\label{eq:10}
\end{equation}

where \( q_{ij} = K_{ij} \widehat{Y}_{ij} + (1 - K_{ij})(1 - \widehat{Y}_{ij}) \), and \(\gamma\) is the focusing parameter of the focal loss. The formula for L1 loss is:
\begin{equation}
	L_{l1} = \frac{1}{H_1 W_1} \sum_{i, j=1}^{H_1, W_1} |K_{ij} - \widehat{Y}_{ij}|
	\label{eq:11}
\end{equation}

where \(K\) is the ground truth mask.

Finally, the total segmentation loss is the sum of the focal loss and L1 loss:
\begin{equation}
	L_{seg} = L_{focal} + L_{l1}
	\label{eq:12}
\end{equation}
\begin{table}[!b]
	\centering
	\caption{Average Image-Level Anomaly Detection AUC (\%) Across All Categories on the MVTec AD Dataset}
	\label{tab:1}
	\resizebox{0.8\textwidth}{!}{
		\begin{tabular}{cccccccc}
			\toprule
			STPM\cite{Wang2021StudentTeacherFP} & DRAEM\cite{zavrtanik2021draem} & DSR\cite{zavrtanik2022dsr} & PatchCore\cite{Roth2021TowardsTR} & DeSTSeg\cite{Zhang2022DeSTSegSG} & CND\cite{wang2025cnc} & Ours \\
			\midrule
			95.1 & 98 & 98.2 & 98.5 & 98.4 & 98.6 & \textbf{98.9} \\
			\bottomrule
		\end{tabular}
	}
\end{table}

\section{Experiments}
\subsection{Dataset}
We comprehensively evaluated our proposed method on the MVTec AD \cite{bergmann2019mvtec} dataset. MVTec AD is a widely recognized benchmark for unsupervised anomaly detection and localization, consisting of a total of 15 categories, including 10 object categories and 5 texture categories. Each category provides a training set of several hundred normal images and a test set containing both normal and anomalous images. Anomalous images in the test set include pixel-precise ground truth annotations for evaluation. Image resolutions range from \(700 \times 700\) to \(1024 \times 1024\) pixels. For the external data source \(A\) used in synthetic anomaly generation (Equation \eqref{eq:1}), we utilized the Describable Textures Dataset (DTD) \cite{cimpoi2014describing}. Although \cite{zavrtanik2021draem} shows that other datasets, such as ImageNet, can achieve comparable performance, the DTD dataset is more compact, easier to use, and better suited for generating anomalies.

\subsection{Evaluation Metrics}\label{4.2}
We used three levels of evaluation metrics for the examination of model's detection performance. For image-level evaluation, we use the Area Under the ROC Curve (AUC), a standard metric in anomaly detection. At the pixel level, we use Average Precision (AP) \cite{saito2015precision}, as it better handles the significant class imbalance typical in anomaly detection datasets compared to AUC. For instance-level evaluation, practical applications (e.g., industrial defect and medical lesion detection) often prioritize locating anomalous instances (entirely or partially) over precise pixel-wise detection. One common metric is the Per-Region Overlap (PRO) score proposed in \cite{bergmann2020uninformed}. PRO assigns equal weight to ground truth connected regions of varying sizes and measures overlap with predictions at a user-defined false positive rate (e.g., \(30\%\)). However, recognizing the critical role of instance recall in real-world applications, we propose Instance Average Precision (IAP) as a more intuitive metric. For IAP, anomalous instances are defined as the largest connected regions in the ground truth. An instance is considered detected if over \(50\%\) of its pixels are correctly classified as positive. By setting different thresholds, we can plot the curve of pixel-level precision versus instance-level recall, and the area under this curve represents the IAP. Furthermore, for scenarios demanding high recall, we also compute precision at an instance recall of \(k\%\) (IAP@\(k\)). In our experiments, \(k=90\) to simulate high-risk applications.
\begin{table}[!t]
	\centering
	\caption{AUC/AP (\%) Results for Pixel-Level Anomaly Localization on the MVTec AD Dataset}
	\label{tab:2}
	\resizebox{0.9\textwidth}{!}{
		\begin{tabular}{@{}c|lccccccc@{}}
			\toprule
			\multicolumn{2}{c}{Category} & STPM\cite{Wang2021StudentTeacherFP} & DRAEM\cite{zavrtanik2021draem} & DSR\cite{zavrtanik2022dsr} & PatchCore\cite{Roth2021TowardsTR} & DeSTSeg\cite{Zhang2022DeSTSegSG} & CND\cite{wang2025cnc} & Ours \\
			\midrule
			\multirow{6}{*}{texture} 
			& carpet & 99.1 / 65.3 & 96.2 / 64.4 & - / \textbf{78.2} & 99.1 / 66.7 & 93.5 / 62.9 & \textbf{99.3} / 70.7 & 96.6 / 75.4 \\
			& grid & 99.1 / 45.4 & \textbf{99.5} / 56.8 & - / \textbf{68.0} & 98.9 / 41.0 & 99.2 / 61.9 & 98.4 / 25.8 & 98.6 / 58.6 \\
			& leather & 99.2 / 42.9 & 98.9 / 69.9 & - / 62.5 & 99.4 / 51.0 & \textbf{99.8} / \textbf{77.3} & 99.5 / 50.1 & 99.7 / 76.4 \\
			& tile & 96.6 / 61.7 & \textbf{99.5} / \textbf{96.9} & - / 93.9 & 96.6 / 59.3 & 98.0 / 94.4 & 97.7 / 73.4 & 98.3 / 91.6 \\
			& wood & 95.2 / 47.0 & 97.0 / 80.5 & - / 68.4 & 95.1 / 52.3 & 96.2 / 71.7 & 96.4 / 63.3 & \textbf{97.8} / \textbf{82.4} \\
			\cline{2-9} \\[-1.5ex]
			& average & 97.8 / 52.5 & \textbf{98.2} / 73.7 & - / 74.2 & 97.8 / 54.1 & 97.3 / 73.6 & \textbf{98.2} / 56.6 & \textbf{98.2} / \textbf{76.9} \\
			\midrule
			\multirow{11}{*}{object} 
			& bottle & 98.8 / 80.6 & \textbf{99.3} / 89.8 & - / \textbf{91.5} & 98.9 / 80.1 & 99.1 / 90.2 & 99.0 / 81.8 & \textbf{99.3} / 91.3 \\
			& cable & 94.8 / 58.0 & 95.4 / 62.6 & - / \textbf{70.4} & \textbf{98.8} / 70.0 & 97.0 / 59.8 & 98.2 / 64.1 & 96.6 / 63.8 \\
			& capsule & 98.2 / 35.9 & 94.1 / 43.5 & - / 53.3 & \textbf{99.1} / 48.1 & 97.6 / 53.3 & 98.2 / 36.9 & 99.0 / \textbf{59.3} \\
			& hazelnut & 98.9 / 60.3 & 99.5 / 88.1 & - / 87.3 & 99.0 / 61.5 & 99.5 / \textbf{89.6} & 98.8 / 53.3 & \textbf{99.6} / 89.3 \\
			& metal nut & 97.2 / 79.3 & 98.7 / 91.7 & - / 67.5 & 98.8 / 88.8 & \textbf{98.9} / 93.4 & 95.5 / 68.4 & \textbf{98.9} / \textbf{94.2} \\
			& pill & 94.7 / 63.3 & 97.6 / 46.1 & - / 65.7 & 98.2 / 78.7 & 98.4 / \textbf{84.9} & \textbf{98.8} / 80.2 & \textbf{98.8} / 84.2 \\
			& screw & 98.6 / 26.9 & \textbf{99.7} / \textbf{71.5} & - / 52.5 & 99.5 / 41.4 & 98.5 / 59.1 & 99.0 / 26.2 & 97.6 / 58.4 \\
			& toothbrush & 98.9 / 48.8 & 98.1 / 54.7 & - / 74.2 & 98.9 / 51.6 & \textbf{99.4} / \textbf{76.9} & 99.0 / 49.7 & \textbf{99.4} / 71.5 \\
			& transistor & 81.9 / 44.4 & 90.0 / 51.7 & - / 41.1 & \textbf{96.2} / 63.2 & 86.1 / \textbf{79.1} & 94.5 / 57.3 & 94.2 / 65.8 \\
			& zipper & 98.0 / 54.9 & 98.6 / 72.3 & - / 78.5 & 99.0 / 64.0 & 99.1 / 83.9 & 97.6 / 45.1 & \textbf{99.2} / \textbf{84.0} \\
			\cline{2-9} \\[-1.5ex]
			& average & 96.0 / 55.2 & 97.2 / 67.1 & - / 68.2 & \textbf{98.6} / 64.8 & 97.4 / \textbf{77.0} & 97.8 / 56.3 & 98.3 / 76.2 \\
			\midrule
			\multicolumn{2}{c}{average} & 96.6 / 54.3 & 97.5 / 69.3 & - / 70.2 & \textbf{98.4} / 61.2 & 97.4 / 75.9 & 98.0 / 56.4 & 98.2 / \textbf{76.4} \\
			\bottomrule
		\end{tabular}
	}
\end{table}

\subsection{Implementation Details}
In the experiment, the teacher network is based on the ResNet18 \cite{he2016deep} model pre-trained on ImageNet \cite{deng2009imagenet}, and the weights of the teacher network remain fixed throughout the training process. The encoder of the denoising student network uses the architecture of PA\_ResNet18, and the decoder also uses the reverse architecture of PA\_ResNet18 (replacing all downsampling with bilinear upsampling). Both the encoder and decoder are constructed by concatenating four residual blocks in an equal manner, with all weights randomly initialized and participating in training. The segmentation network consists of two improved residual blocks (PA\_Residual Blocks) and a rectangular self-calibration module (RCM), all of which are also randomly initialized and participate in the training process. The denoising student network and the segmentation network were trained using stochastic gradient descent (SGD). The learning rate of the denoised student network is set to 0.5, the learning rate of the improved residual block in the segmentation network is set to 0.1, the learning rate of the RCM module is set to 0.01, and the batch size is 16. During training, the denoised student network is first trained for 3000 iterations, after which its weights are fixed. The segmentation network then undergoes an additional 4000 iterations of training. All input images were resized to \(256 \times 256\) during training and testing.

\subsection{Results}
We compare our method with several leading contemporary approaches using the evaluation metrics detailed in Sec.~\ref{4.2}. Unavailable experimental results from other methods are marked as "-".
\begin{table}[t]
	\centering
	\caption{Instance-Level Anomaly Detection IAP/IAP@90 (\%) Results on the MVTec AD Dataset}
	\label{tab:3}
	\resizebox{0.8\textwidth}{!}{
		\begin{tabular}{@{}c|lccccc@{}}
			\toprule
			\multicolumn{2}{c}{Category} & STPM\cite{Wang2021StudentTeacherFP} & DRAEM\cite{zavrtanik2021draem} & PatchCore\cite{Roth2021TowardsTR} & DeSTSeg\cite{Zhang2022DeSTSegSG} & Ours \\
			\midrule
			\multirow{6}{*}{texture} 
			& carpet & 68.4 / 52.2 & 76.8 / 32.3 & 64.4 / 43.7 & 76.0 / 30.8 & \textbf{87.6} / \textbf{70.4} \\
			& grid & 45.7 / 21.0 & 55.5 / 42.3 & 39.1 / 15.6 & \textbf{62.2} / \textbf{45.1} & 57.8 / 30.2 \\
			& leather & 46.2 / 24.9 & 78.6 / 55.0 & 50.1 / 30.1 & 78.1 / 67.3 & \textbf{81.2} / \textbf{70.4} \\
			& tile & 62.9 / 55.3 & \textbf{98.9} / \textbf{98.2} & 60.0 / 52.1 & 97.4 / 93.9 & 96.2 / 92.6 \\
			& wood & 56.0 / 35.4 & 88.4 / 72.6 & 59.7 / 35.6 & 73.3 / 57.0 & \textbf{89.5} / \textbf{82.8} \\
			\cline{2-7} \\[-1.5ex]
			& average & 55.8 / 37.8 & 79.6 / 60.1 & 54.7 / 35.4 & 77.4 / 58.8 & \textbf{82.5} / \textbf{69.3} \\
			\midrule
			\multirow{11}{*}{object} 
			& bottle & 83.2 / 73.3 & 90.3 / 84.8 & 81.8 / 70.1 & 91.7 / \textbf{87.4} & \textbf{92.9} / 86.9 \\
			& cable & 54.9 / 17.2 & 47.0 / 10.8 & \textbf{69.2} / \textbf{50.6} & 51.3 / 30.0 & 60.2 / 34.9 \\
			& capsule & 37.2 / 17.9 & 50.7 / 21.4 & 44.2 / 26.9 & 44.6 / 23.5 & \textbf{57.5} / \textbf{40.3} \\
			& hazelnut & 64.8 / 56.2 & \textbf{95.7} / \textbf{89.0} & 63.8 / 52.5 & 88.1 / 79.6 & 86.9 / 76.5 \\
			& metal nut & 83.4 / 81.7 & 92.6 / 83.9 & 90.1 / 84.6 & 92.3 / 83.1 & \textbf{95.0} / \textbf{88.3} \\
			& pill& 72.0 / 45.5 & 46.9 / 41.5 & 82.7 / 63.5 & 77.8 / 38.9 & \textbf{88.1} / \textbf{74.9} \\
			& screw & 24.4 / 4.2 & \textbf{68.8} / \textbf{33.0} & 38.4 / 16.3 & 53.8 / 6.4 & 52.7 / 5.3 \\
			& toothbrush & 41.9 / 23.4 & 44.7 / 21.5 & 40.4 / 22.1 & 60.6 / 42.9 & \textbf{62.8} / \textbf{45.9} \\
			& transistor & 53.4 / 8.5 & 59.3 / 22.8 & 69.9 / 36.8 & \textbf{83.8} / \textbf{69.9} & 82.6 / 65.1 \\
			& zipper & 59.1 / 46.6 & 78.7 / 67.0 & 66.0 / 52.4 & 88.5 / 74.5 & \textbf{89.0 } / \textbf{76.7} \\
			\cline{2-7} \\[-1.5ex]
			& average & 57.5 / 37.4 & 67.5 / 47.5 & 64.6 / 47.6 & 73.2 / 53.6 & \textbf{76.8} / \textbf{59.5} \\
			\midrule
			\multicolumn{2}{c}{average} & 56.9 / 37.5 & 71.5 / 51.7 & 61.3 / 43.5 & 74.6 / 55.4 & \textbf{78.7} / \textbf{62.7} \\
			\bottomrule
		\end{tabular}
	}
\end{table}

For the image-level anomaly detection, we present the AUC scores of our method and other methods in the image-level anomaly detection task in Table~\ref{tab:1}, with a focus on the average results across the 15 categories of the MVTec AD dataset. As observed, our method outperforms others, achieving an Image\_AUC of \(98.9\%\). In Table~\ref{tab:2}, we report the AUC and AP metrics for the pixel-level anomaly localization task. Our method achieves nearly identical AUC scores compared to PatchCore \cite{Roth2021TowardsTR}, while setting a new benchmark in terms of AP. Additionally, our approach demonstrates substantial improvements across most categories, with scores either at or near the highest, strongly suggesting that it performs robustly across a wide range of categories.

Instance-level anomaly detection results are presented in Table~\ref{tab:3}, featuring the IAP and IAP@90 metrics for comprehensive assessment. The results show that our method achieves highly competitive performance on both metrics, demonstrating strong anomaly detection capabilities. Notably, our method reaches an average IAP@90 of \(62.7\%\). This means that when the model successfully detects \(90\%\) of anomalous instances, the pixel-level precision is \(62.7\%\). This translates to a pixel-level false positive rate of only \(37.3\%\) under these conditions, suggesting a good balance of recall and precision, indicative of robustness for real-world applications. The achieved precision at high instance recall suggests our method's potential utility in demanding industrial environments and other critical areas where reliable anomaly detection is paramount.

\begin{figure}
	\centering
	\includegraphics[width=0.8\textwidth]{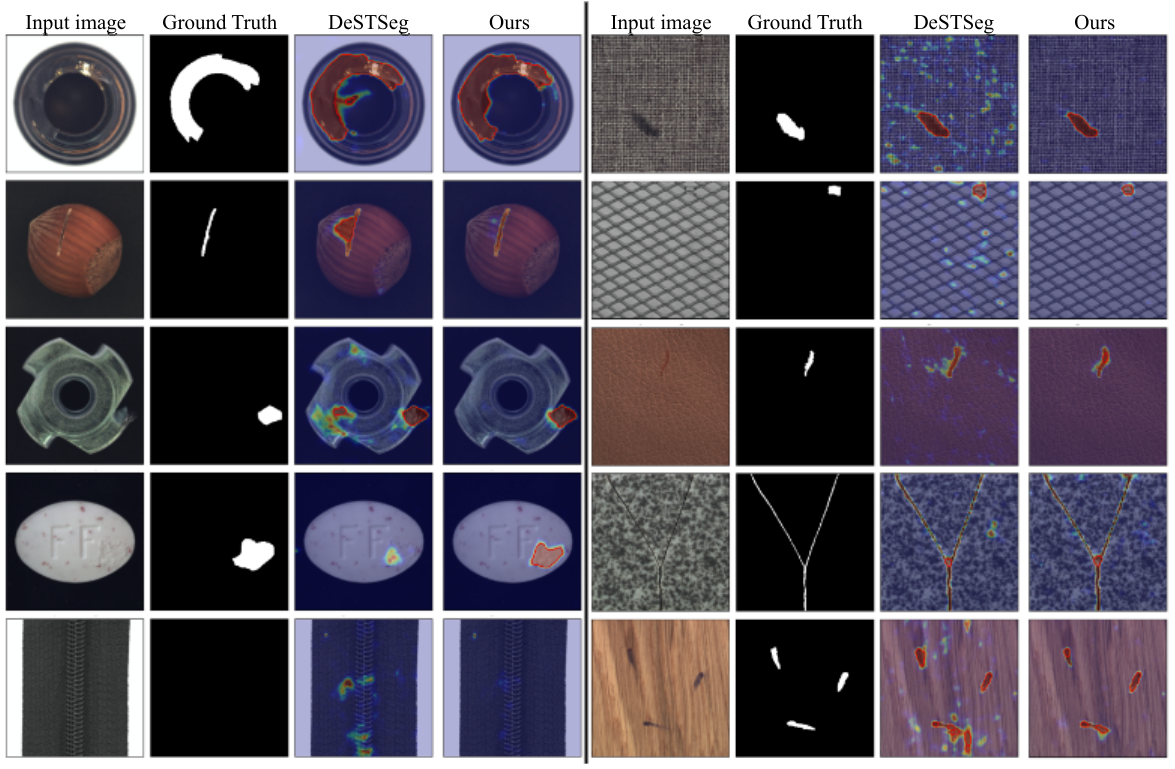}
	\caption{Visual comparison of anomaly segmentation between our method and DeSTSeg. For each example, first column: input image; second column: Ground Truth; third and fourth columns: predicted segmentation results.}
	\label{fig:Visualization}
\end{figure}

Additionally, Fig.~\ref{fig:Visualization} provides a visual comparison of our method against the DeSTSeg approach \cite{Zhang2022DeSTSegSG}. The qualitative results indicate that our method produces segmentation masks with improved visual quality compared to DeSTSeg. Particularly for the five texture categories (shown on the right), our method demonstrates a better capability to accurately highlight anomalous regions while effectively suppressing irrelevant noise.

\subsection{Ablation Studies}
In Table~\ref{tab:4}, we conducted ablation experiments to effectively evaluate the proposed three modules. Specifically, we adopted the network architecture from \cite{Zhang2022DeSTSegSG} as the baseline and performed a detailed analysis and comparative experiments to assess the independent effects of each module as well as their combined impact on network performance. The experimental results demonstrate that, by comparing Experiment 2 with 5 and Experiment 3 with 6, integrating the proposed PCAR module with either the RCM or AFF module significantly enhances the network's performance. Furthermore, the results of Experiment 7 further reveal that incorporating the PCAR, RCM, and AFF modules into the network architecture achieves optimal performance. This combination not only enables more accurate detection and localization but also substantially improves the overall network effectiveness, thereby validating the efficiency and superiority of our proposed approach.
\begin{table}
	\centering
	\caption{Ablation study on the modules: The results of RCM, AFF, and PCAR individually and in combination, evaluated on the MVTec AD dataset, for image-level AUC (\%), pixel-level AP (\%), and instance-level IAP (\%).}
	\label{tab:4}
	\resizebox{0.6\textwidth}{!}{
		\begin{tabular}{ccccccc}
			\toprule
			& RCM & AFF & PCAR & img(AUC) & pix(AP) & ins(IAP) \\
			\midrule
			1 &  &  &  & 98.4 & 75.9 & 74.6 \\
			2 & \checkmark &  &  & 98.8 & 74.0 & 76.6 \\
			3 &  & \checkmark &  & 97.9 & 74.2 & 74.5 \\
			4 &  &  & \checkmark & 98.8 & 76.2 & 78.1 \\
			5 &\checkmark &  & \checkmark & 98.1 & \textbf{76.4} & 78.4 \\
			6 &  & \checkmark & \checkmark & \textbf{98.9} & 75.3 & 76.5 \\
			7 & \checkmark & \checkmark & \checkmark & \textbf{98.9} & \textbf{76.4} & \textbf{78.7} \\
			\bottomrule
		\end{tabular}
	}
\end{table}

\section{Conclusion}
In this paper, we propose a novel PCAR module and integrate it into our PFADSeg model. The PCAR module employs parallel convolution techniques to effectively capture multi-scale spatial information, enabling precise localization and segmentation of anomalous regions while suppressing irrelevant noise in detection images. The PFADSeg model further optimizes the feature extraction process between the student and teacher networks, using an improved strategy to enhance discriminative features in anomalous regions. Additionally, a segmentation network is integrated to adaptively fuse features from the student-teacher network, achieving more accurate anomaly segmentation. Experiments on the MVTec AD dataset demonstrate that our method outperforms state-of-the-art approaches. Specifically, in image-level anomaly detection, our method achieves an AUC of \(98.9\%\), surpassing the current best by \(0.3\%\). For pixel-level anomaly localization, it achieves an AP of \(76.4\%\), exceeding the best by \(0.5\%\), with AUC nearing the state-of-the-art. Notably, for instance-level anomaly detection, our method improves IAP@\(90\) by \(7.3\%\) and IAP by \(4.1\%\). These results validate the effectiveness and superiority of our approach.

\section*{Acknowledgments}
This preprint has not undergone peer review (when applicable) or any post-submission improvements or corrections. The Version of Record of this contribution is published in \textit{Neural Information Processing}, and is available online at \url{https://doi.org/10.1007/978-981-95-4109-6_17}.
%
%
%
%
%
\bibliographystyle{splncs04}
\bibliography{reference}
\end{document}